\documentclass[letterpaper, 10 pt, conference]{ieeeconf}  

\IEEEoverridecommandlockouts                              
\usepackage{graphics} 
\usepackage{epsfig} 
\usepackage{mathptmx} 
\usepackage{times} 
\usepackage{amssymb}  
\usepackage{mathtools}
\usepackage{dblfloatfix}
\usepackage{color}
\usepackage{booktabs,tabularx}
\usepackage{optidef}
\usepackage[font=small,skip=3pt]{caption}
\usepackage{color,soul}

\usepackage{siunitx}
\usepackage[style=ieee,,doi=false,url=false,eprint=false,isbn=false]{biblatex}

\defbibenvironment{bibliography}
  {\list
     {\printtext[labelnumberwidth]{%
        \printfield{prefixnumber}%
        \printfield{labelnumber}}}
     {\setlength{\labelwidth}{\labelnumberwidth}%
      \setlength{\leftmargin}{0.8\labelwidth}%
      \setlength{\labelsep}{0.5\biblabelsep}%
      \addtolength{\leftmargin}{\labelsep}
      \setlength{\itemsep}{\bibitemsep}%
      \setlength{\parsep}{\bibparsep}}%
      }
  {\endlist}
  {\item}

\usepackage{subfigure}
\renewcommand{\thefigure}{\arabic{figure}}
\title{\LARGE \bf
Hybrid Soft Electrostatic Metamaterial Gripper for Multi-surface, Multi-object Adaptation
}

\author{Ryo Kanno$^{1,2}$, \textit{Student Member, IEEE}, Pham H. Nguyen$^{1,3,*}$, \textit{Member, IEEE}, Joshua Pinskier$^{4}$, \textit{Member, IEEE},\\ David Howard$^{4}$, \textit{Member, IEEE}, Sukho Song$^{1,3}$, \textit{Member, IEEE}, Mirko Kovac$^{1,2,3,*}$, \textit{Member, IEEE}
\thanks{R. Kanno, P. H. Nguyen, S. Song, and M. Kovac are with the Laboratory of Sustainability Robotics, Empa - Swiss Federal Laboratories for Materials Science and Technology, 8600 Dübendorf, Switzerland. {\tt\small $\{$ryo.kanno, huy.pham, sukho.song, mirko.kovac$\}$@empa.ch}}
\thanks{R. Kanno, and M. Kovac are with the École Polytechnique Fédérale de Lausanne, 1005 Lausanne, Switzerland.}
\thanks{P. H. Nguyen, S. Song, and M. Kovac are with the Aerial Robotics Laboratory, Imperial College London, South Kensington Campus, London, SW7 2AZ, United Kingdom.}
\thanks{J. Pinskier and D. Howard are with the Commonwealth Scientific and Industrial Research Organisation (CSIRO). {\tt\small $\{$josh.pinskier, david.howard$\}$@data61.csiro.au}
\thanks{$*$ Address all correspondence to these authors.}}
}

\begin{document}

\maketitle
\thispagestyle{empty}
\pagestyle{empty}

\begin{abstract}

One of the trendsetting themes in soft robotics has been the goal of developing the ultimate universal soft robotic gripper. One that is capable of manipulating items of various shapes, sizes, thicknesses, textures, and weights. All the while still being lightweight and scalable in order to adapt to use cases. In this work, we report a soft gripper that enables delicate and precise grasps of fragile, deformable, and flexible objects but also excels in lifting heavy objects of up to $1617x$ its own body weight. The principle behind the soft gripper is based on extending the capabilities of electroadhesion soft grippers through the enhancement principles found in metamaterial adhesion cut and patterning. This design amplifies the adhesion and grasping payload in one direction while reducing the adhesion capabilities in the other direction. This counteracts the residual forces during peeling (a common problem with electroadhesive grippers), thus increasing its speed of release. In essence, we are able to tune the maximum strength and peeling speed, beyond the capabilities of previous electroadhesive grippers. We study the capabilities of the system through a wide range of experiments with single and multiple-fingered peel tests. We also demonstrate its modular and adaptive capabilities in the real-world with a two-finger gripper, by performing grasping tests of up to $5$ different multi-surfaced objects. 
\end{abstract}
\section{INTRODUCTION}
\begin{figure}[t]
\centering
\includegraphics[width=0.46\textwidth]{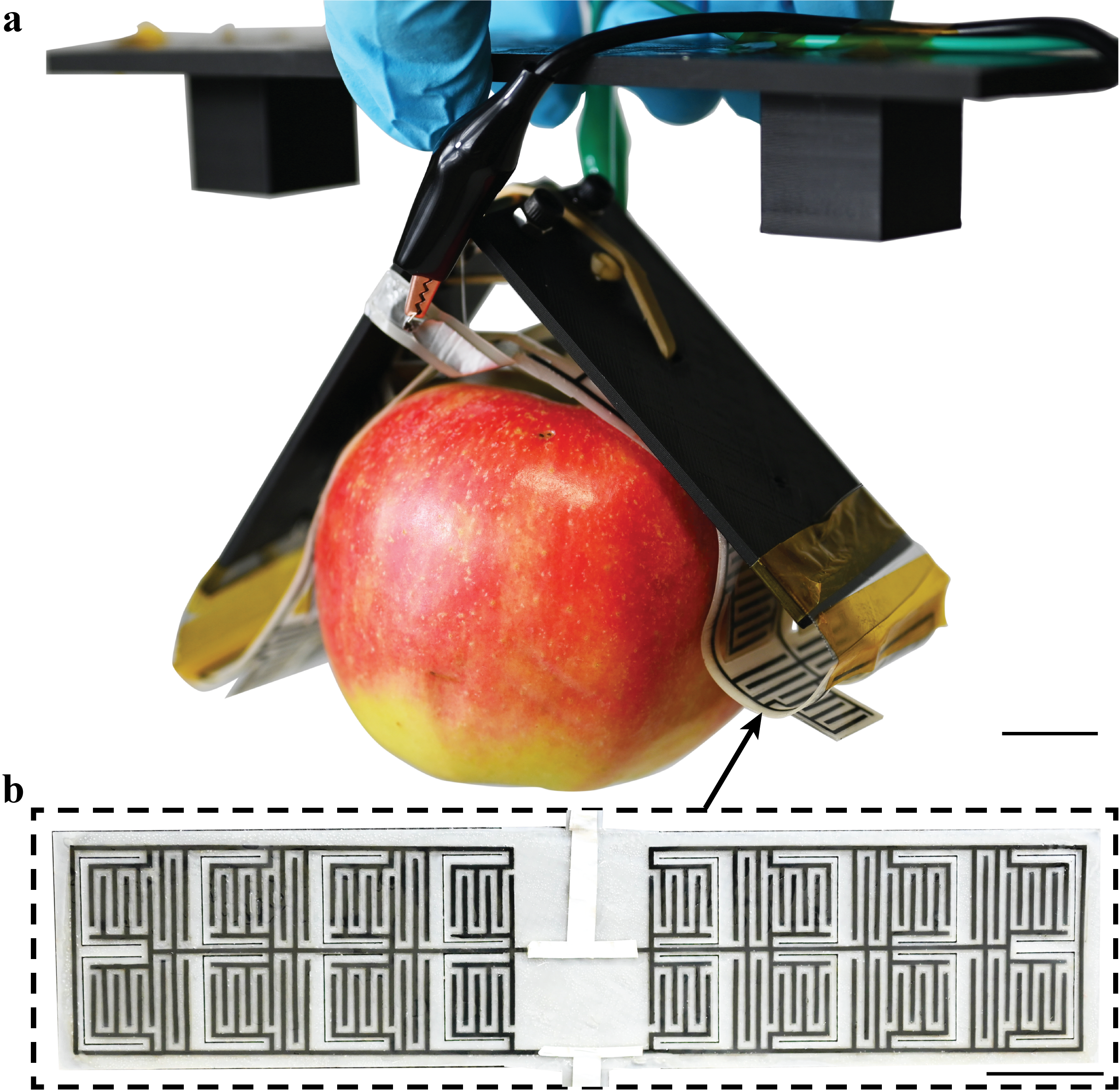}
\setlength{\belowcaptionskip}{-10pt}
\caption{(a) Soft Electrostatic Metamaterial (SEM) gripper. (b) The SEM adhesive. The scale bar is 20 mm.}
\label{fig:fig1}
\vspace{-1em}
\end{figure}
\begin{figure*}[t!]
\centering
\includegraphics[width=0.98\textwidth]{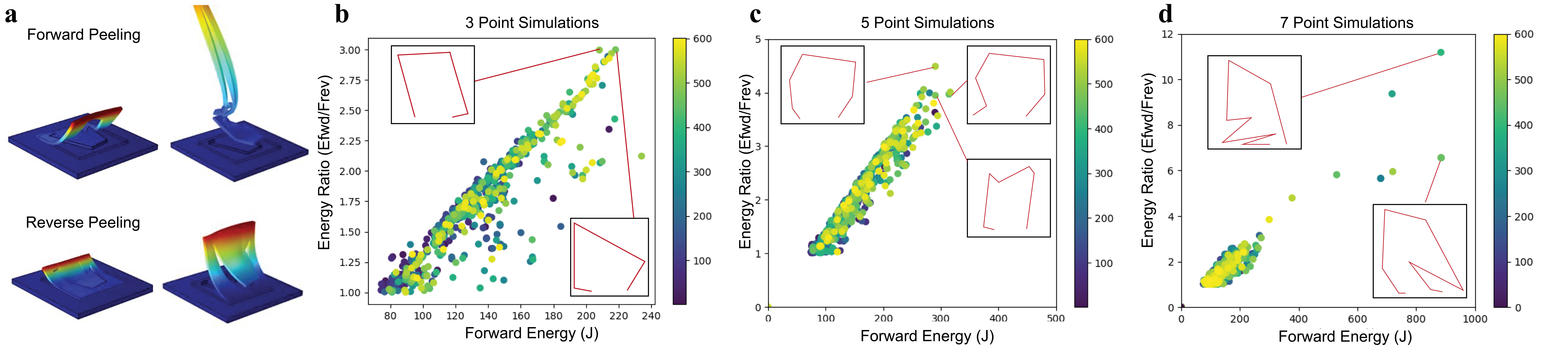}
\caption{Adhesion Simulation in COMSOL. a) The metamaterial is first evaluated by lifting in the forward direction until the material releases the substrate [Top], then repeated in the reverse direction [Bottom]. Results of (b) 3 Point Polygon, (c) 5 Point Polygon, and (d) 7 Point Polygon optimizations, showing high-performing shapes and optimization chronology. The color bar shows the sequence of design evaluations, with brighter colors occurring later. The clustering of bright numbers at the top right is indicative of optimizer progress.}
\label{fig:Sim_peeling}
\vspace{-1.5em}
\end{figure*}
The race towards the ultimate universal soft gripper has been on display since the dawn of soft robotics~\cite{shintake2017}. The goal has always been to develop a lightweight, compliant, robust, and versatile system that is capable of diverse grasps ranging from handling delicate, flat, and deformable objects with varying weights, material properties, surface textures, shapes and sizes~\cite{hughes2016}. To achieve this balance between various trade offs such as innate softness, precision, and grasp strength, electroadhesive gripper technology has emerged with varying successes and limitations~\cite{shintake2016versatile,guo2019electroadhesion,piskarev2023soft,okuno2019stretchable}. Electrostatic soft grippers have highlighted their capabilities to gently grasp various types of objects, whether that be thin sheets, deformable objects, or fragile objects~\cite{han2020hybrid,shintake2016versatile}. One of their key advantages is their ability to adhere to varying surface roughness. The electrostatic force generated also assists with alignment and engagement to the contact surface of the object. They also highlight the capabilities of self-sensing. One weakness of these electrostatic soft grippers is that their maximum normal stress is quite low, thus limiting their ability to carry heavy loads~\cite{han2020hybrid,shintake2016versatile}. Furthermore, during disengaging from the target object, small residual forces remain, thus slowing down the release phase~\cite{caucciolo2022_peeling}. This might create a challenge when trying to rapidly pick and place delicate objects. To tackle some of the weakness seen with electrostatic soft grippers, researchers have looked at hybrid systems, whether that be combining electroadhesion with suction cups~\cite{okuno2019stretchable}, microfibrillar structures~\cite{alizadehyazdi2020electrostatic}, jamming structures~\cite{piskarev2023soft}, and magnetic augmentation~\cite{guo2019_mag_electro,han2020hybrid}.

In this work, we aim to develop a novel soft gripper, called the Soft Electrostatic Metamaterial (SEM) gripper, that tackles limitations of the electrostatic soft grippers, as seen in Fig.~\ref{fig:fig1}a. In essence, we tackle the problem of residual forces during disengaging the gripper, by controlling the directionality and force of peeling. At the same time, we also amplify the lifting capacity of the gripper.  

In order to achieve this, we integrate metamaterial cuts with the localized dielectric elastomer substrate, as seen in Fig.~\ref{fig:fig1}b. The metamaterial cuts are inspired by the capabilities of metamaterial adhesives seen in Hwang et. al~\cite{Hwang2023_metamaterial}. To our knowledge, metamaterial adhesion has not been utilized in soft robotic grippers, because they are still reliant on the substrate material for their adhesive properties and require mechanical preloading onto the target object to be effective.  

With the combination of metamaterial cuts and electroadhesion, we are able to tune the electroadhesion force to pull the SEM membrane into contact with the object’s surface, and provide the preloading required to fully engage the adhesive, without unpredictable deformation or damage on the target object’s surface. The combination of these two technologies led to the introduction of the SEM gripper that can handle flat, irregularly-shaped, and oversized objects with different textures in one system.

The remainder of the paper is as follows. Section~\ref{sec:act_design_fab} introduces the working principle, shape optimization, and fabrication of the SEM adhesives. Section~\ref{sec:characterization and Optimization} presents characterizations of the adhesive, utilizing the cut patterns determined through simulation. Section~\ref{sec:gripper_charac} highlights the grasping force characterization and pick-and-release demonstration of the adhesives and the SEM gripper’s graspability with up to $5$ different surfaces of objects. Section~\ref{sec:conclude} concludes this paper and discusses future directions.


\section{Design, Shape Optimization, and Fabrication}
\label{sec:act_design_fab}
The SEM gripper was developed to improve three main limitations seen with electrostatic soft grippers and metamaterial adhesives. First, was the reduction of the residual forces created by electrostatic grippers during disengagement. Second, was amplifying the electrostatic gripper's grasping force. And finally, was the providing controlled preloading required for metamaterial adhesives.
\begin{table}[b]
    \centering
        \setlength{\belowcaptionskip}{10pt}
    \begin{tabular}{c|c|c}
        Design & \shortstack{Forward\\ Energy} &  \shortstack{Energy\\Ratio} \\
        \hline
         Square & 154.21 & 1.58 \\
         Triangle & 278.21 & 4.06 \\
         Hybrid & 205.73 & 2.54\\
    \end{tabular}
    \caption{Reference Shape Simulation Results}
    \label{tab:reference_shape_sim}
    \vspace{-1.5em}
\end{table}
\subsection{Working Principle}
To design and develop this gripper, we utilized two heritage projects as both reference and comparison~\cite{shintake2016versatile,Hwang2023_metamaterial}. In Fig.~\ref{fig:fig1}b, we notice the combination of interdigitated electrodes and metamaterial cut patterns. Through the programmable metamaterial cut architecture, we are able to achieve amplified adhesion in one direction, while releasable adhesion in the other. Thus overcoming the residual force during disengagment in one direction and improving the overall grasping force. Further, by activating electrodes, we are able to create electrostatic force to engage the adhesive to the surface of the grasping object to activate the metamaterial adhesive properly, as seen in Fig.~\ref{fig:gripper_mechanism}c. 

\begin{figure*}[t!]
\centering
\includegraphics[width=0.94\textwidth]{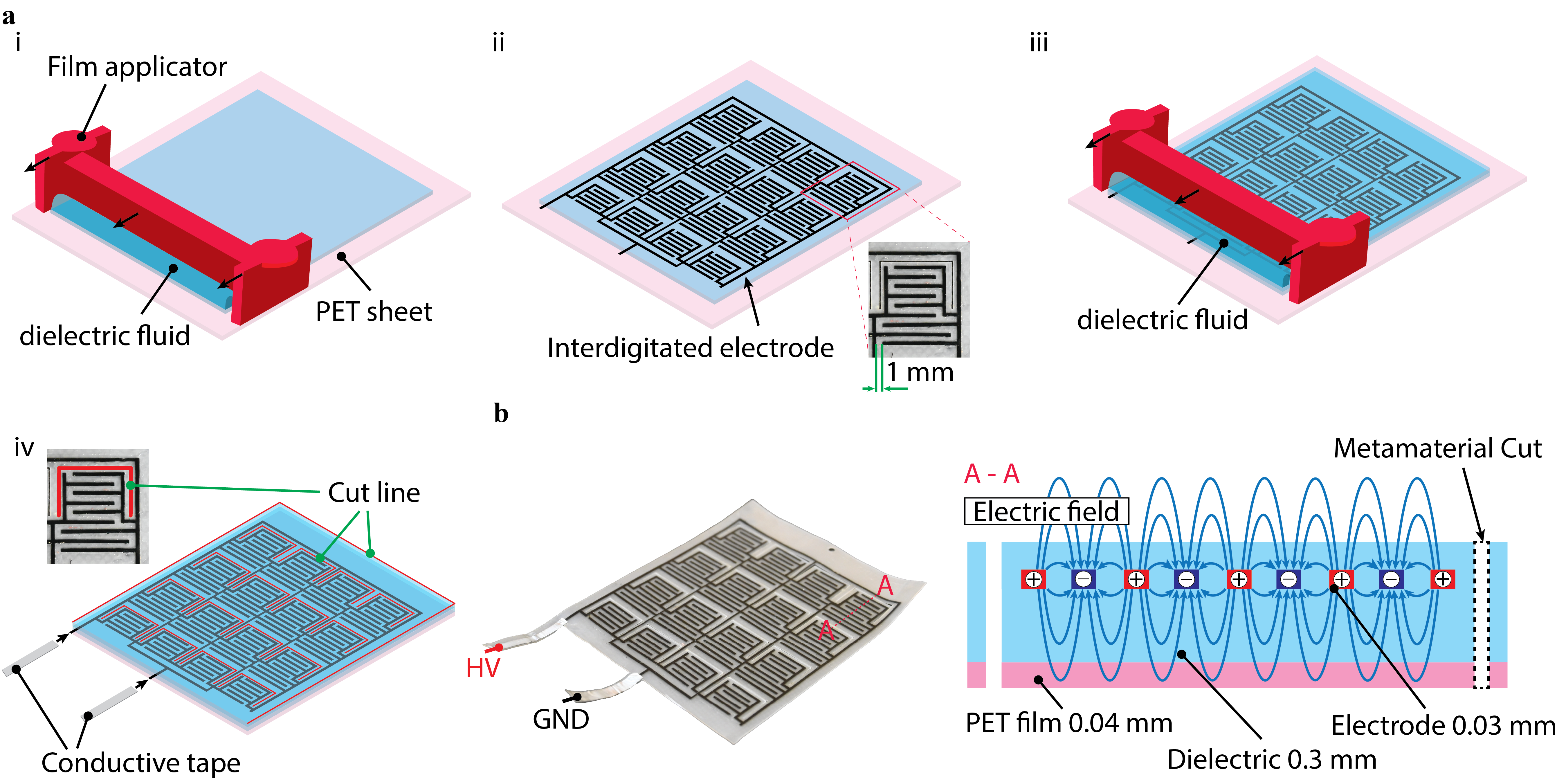}
\caption{Fabrication process. a-i) First dielectric mixture layer is blade-casted by a film applicator. a-ii) Laser cut interdigitated electrodes are attached to the pre-cured dielectric layer onto a PET sheet. a-iii) Second dielectric mixture layer is blade-casted. a-iv) Metamaterial cut lines are created using laser cutter. Conductive tapes are attached. b) Prototyped SEM adhesive [Left]. An electrostatic adhesion is generated by applying high voltage. [Right]}
\label{fig:fabrication}
\vspace{-1.5em}
\end{figure*}

\subsection{Shape Optimization}
Efficient computational design tools are often required to explore the large space of possible designs, which exploit available data and rapidly iterate over the design space \cite{https://doi.org/10.1002/aisy.202100086}.
Here, the shape of the metamaterial cut pattern will significantly impact adhesion performance, hence a series of model-based grasp optimizations were undertaken to identify high-performing cut geometries.

A multi-objective optimization problem was established to find designs which give both high-adhesion force and high adhesion directionality, which corresponds to low detachment force, and solved using the well known NSGA-II evolutionary algorithm \cite{996017}. As forces give very noise measurements, internal energy is used to assess adhesion force. 
The cut geometries are parameterized by a set of open-ended polygons with fixed endpoints and $n$ internal points where $n \in \{3,5,7\}$. Hence, three sets of optimizations are performed, each with $2n$ optimization variables, the $x$ and $y$ coordinate for each point. In each optimization run, 600 designs are evaluated, comprising 12 generations each with a population of 50. Some are highlighted in Fig.~\ref{fig:Sim_peeling}.

Polygons are formed by joining points in the clockwise direction and swapping intersecting lines as in Auer et. al \cite{Auer1996HeuristicsFT}. Designs are evaluated using an adhesion study in COMSOL multiphysics, with shear and tensile strength $\SI{80}{\mega\pascal}$, energy release rate $\SI{2.8}{\kilo\joule\per\meter\squared}$, and Mode Mixity exponent $2.284$. In each simulation, the metamaterial sheet is initially attached to an acrylic substrate, and raised from one side in $\SI{0.1}{\milli\meter}$ increments until detachment. The same process is then repeated with the displacement applied at the opposite end to find the reverse direction release energy and calculate the ratio of forward to reverse energy.
This process is illustrated in Fig.~\ref{fig:Sim_peeling}a.


Three benchmark designs are used for comparison: square, triangular and hybrid (house-shaped) designs. Data from these reference shapes are presented in Table \ref{tab:reference_shape_sim} for comparison with the 3 sets of optimizations. 

The results from the 3 optimizations are presented in Fig.\ref{fig:Sim_peeling}b-d, illustrating the high performing designs. The presented results exclude several outlying points with nominal performance several orders of magnitude better than those presented. 
These non-physical results occur in designs with very high aspect ratio sections (i.e long, thin parts).
From the simulated results, three features are noticeable: firstly, it appears to be advantageous to occupy a large area, and have a pointed upper edge. Secondly that there is a strong correlation between forward grasp strength and grasp ratio, this occurs because the reverse direction release energy is relatively insensitive to geometry, and varies within a smaller range than the forward direction.



\begin{figure*}
\centering
\includegraphics[width=0.94\textwidth]{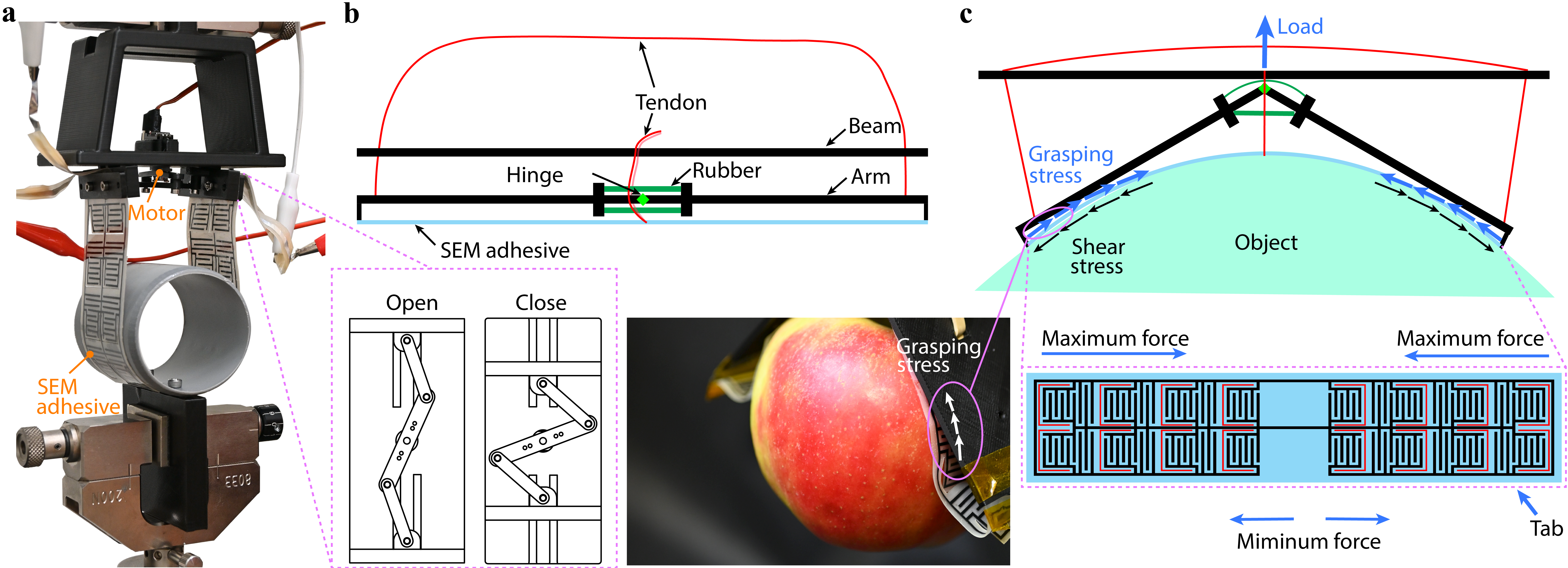}
\caption{SEM gripper Mechanism. a) A parallel gripper controlled with a single servo motor [Top]. The mechanism of the parallel gripper [Bottom]. b) The bistable gripper mechanism consists of two arms connected by a hinge joint. Two rubberbands are connected at the top and bottom of the arm in order to create the bistable mechanism. The tendon assists in pulling it back to its initial state. c) The gripper snaps onto the object as soon as on contact with object, voltage is then applied to enhance adhesion between the object and the adhesive [Top]. The two-arm symmetrical SEM adhesive. Note that the grasping stress applied to SEM adhesive is the same direction for maximum peeling force enduring large payload [Bottom].}
\label{fig:gripper_mechanism}
\vspace{-1.5em}
\end{figure*}

\subsection{Adhesive Fabrication Steps}
To fabricate the SEM adhesives we utilized a layer-by-layer process as seen in Fig.~\ref{fig:fabrication}. The elastomer mixtures in this work were mixed utilizing a planetary centrifugal mixer (ARE-250, Thinky, Japan). First, the urethane mixture (Vytaflex 20, Smooth-on, PA), was blade-casted to a thickness of 200 $\mu$m using a universal film applicator (ZUA 2000 Zehntner, Screening Eagle Technologies, TX) on a 42 $\mu$m thick PET sheet (L480.385, Rausch Packing, Switzerland), highlighted in Fig.~\ref{fig:fabrication}a-i). This layer was pre-cured at room temperature for 20 minutes to obtain a dielectric layer. A 30 $\mu$m carbon black-based double-sided adhesive sheet (ARclad 8006, Adhesive Research, PA), was laser cut (Nova 24, Thunder Laser, Germany) to the interdigitated geometry, as seen in Fig.~\ref{fig:fabrication}a-ii). Note that the line and gap thickness were set at 1 mm in this work. To attach the laser-cut electrode to the urethane film in the previous step, the back end of the electrode is affixed onto an inverted mask of double-sided VHB tape (VHB4905, 3M, MN). This combination is then attached to the urethane film and peeled, to leave the exposed interdigitated electrode, as shown in Fig.~\ref{fig:fabrication}a.ii). The electrode sheet is then cured at room temperature for three hours. A second layer of urethane or PDMS mixture (LSR4130, Elkem Silicones, Norway) is then blade-casted at 100 $\mu$m thickness, to cover the exposed electrode, as seen in Fig.~\ref{fig:fabrication}a-iii), making the overall thickness of the sheet 300 $\mu$m. The sheet is laser cut to generate the metamaterial geometry, after another 3 hour room temperature cure, seen in Fig.~\ref{fig:fabrication}a-iv). Conductive tapes are attached to the exposed electrode lines, as seen in Fig.~\ref{fig:fabrication}a-iv).

For our characterization tests, we utilized a 4x4 grid, of 10 mm x 15 mm rectangles, as in Fig.~\ref{fig:fabrication}b, comparative to the samples in Hwang et. al~\cite{Hwang2023_metamaterial}. For our gripper characterization tests, similar to Shintake et. al~\cite{shintake2016versatile}, 2x4 grids of the same size were made, as seen in Fig.~\ref{fig:fig1}b.

\begin{figure}[b!]
\centering
\setlength{\belowcaptionskip}{2pt}
\includegraphics[width=0.45\textwidth]{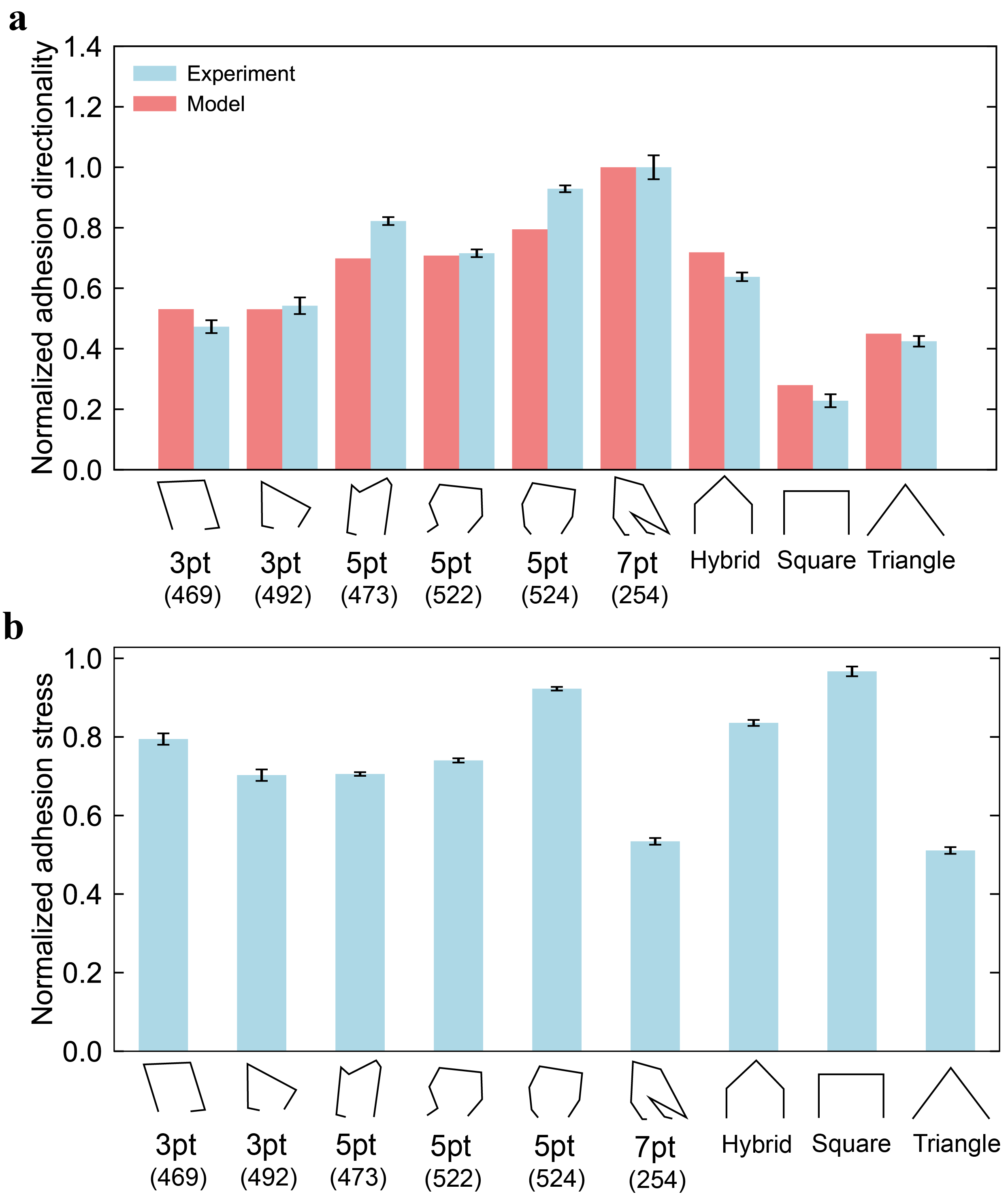}
\caption{a) Normalized adhesion directionality of simulation vs experiment. b) Experimental normalized forward adhesion stress of different shapes.}
\label{fig:Sim_vs_Real}
\end{figure}
The obtained adhesive is shown in Fig.~\ref{fig:fabrication}a-v), with a weight of $1.8$ g. Within the square geometry, electrodes are polarized when high voltage is applied and create a fringe electric field allowing to adhere objects shown in Fig.~\ref{fig:fabrication}b. The adhesives were able to withstand up to 6 kV$/$mm applied by a high-voltage amplifier (CB101, XP Power, NJ).


\subsection{SEM Gripper Design}

In this work, we developed two different types of SEM grippers as seen in Fig.~\ref{fig:gripper_mechanism}. The parallel motion gripper shown in Fig.~\ref{fig:gripper_mechanism}a, employs a motorized mechanism that opens and closes along the same axis. This simple single motor mechanism assists with precision for the testing procedures seen in the next section. The motorized mechanism allows for precise control to set an ideal position, ensuring that the SEM adhesive can be securely attached to the object without causing damage or slippage. 

\begin{figure*}[t]
\centering
\includegraphics[width=0.92\textwidth]{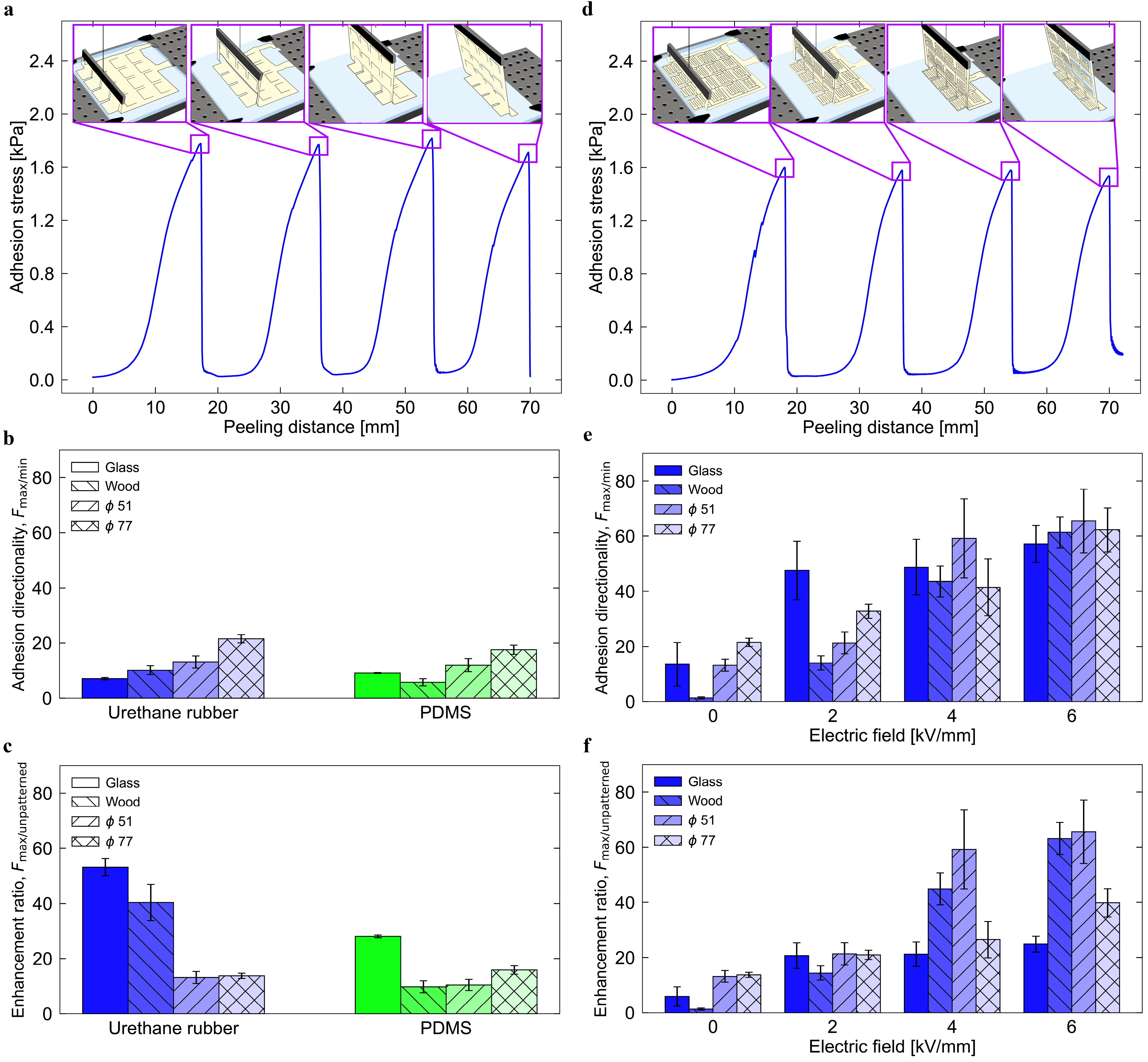}
\caption{Metamaterial vs SEM characterization. Metamaterial adhesion of a) adhesion stress versus peeling distance with roller preloading, b-c) adhesion directionality and enhancement ratio of urethane and PDMS contact layer. Metamaterial and electrostatic co-adhesion of d) adhesion stress versus peeling distance applying 4 kV$/$mm of electric field without roller preloading, e-f) adhesion directionality and enhancement ratio of urethane rubber applying 0, 2, 4, and 6 kV$/$mm of electric field.}
\label{fig:adhesion characterization}
\vspace{-1.5em}
\end{figure*}
On the other hand, the gripper, illustrated in Fig.~\ref{fig:gripper_mechanism}b,c, snaps around a hinge joint. This bistable gripper was inspired by Hawkes et. al~\cite{hawkes2017grasping} that utilized cardboard. In this work, we instead utilized 3D-printed joints, the arms of the gripper utilizing PETG (Bambu Labs, TX), and pre-strained elastic bands to generate bistability. When the object comes into contact with the double-sided SEM adhesive membrane, as seen in  Fig.~\ref{fig:gripper_mechanism}b, the gripper collapses, allowing the SEM adhesive to wrap around the object. At the same time, the SEM gripper is activated (electric field is turned on to approx. 0-4 kV$/$mm), to enhance the gripper-to-object adhesion area. The direction of grasping stress and maximum force is preprogrammed based on the metamaterial cut layout. Note that for flat objects, the tabs are inverted compared to when grasping round objects. When releasing the object, the electroadhesion is turned off and the tendon is pulled. The directionality of the metamaterial adhesion allows quick peeling overcoming the problem of residual electroadhesion holding onto the object. Thus, allowing the gripper to quickly release the object.

\section{SEM Adhesive Characterization}
\label{sec:characterization and Optimization}
In this section, we validated the simulated metamaterial cut models through a series of experimental tests. Our objective was to characterize the adhesive properties using predefined metamaterial cuts configured in a 4x4 grid, applicable to both the metamaterial adhesive and the SEM adhesive, in scenarios with and without roller preload. In addition, testing was done at different flat surface conditions (glass and wood) for both adhesives in order to investigate the difference in adhesion properties. Furthermore, we characterized two different curved surface diameters ($\phi$ = 51 and 77 mm) of PVC pipes. The adhesives were fabricated using either urethane or PDMS, and both materials were characterized as part of this study. 

\begin{figure}[]
\centering
\includegraphics[width=0.44\textwidth]{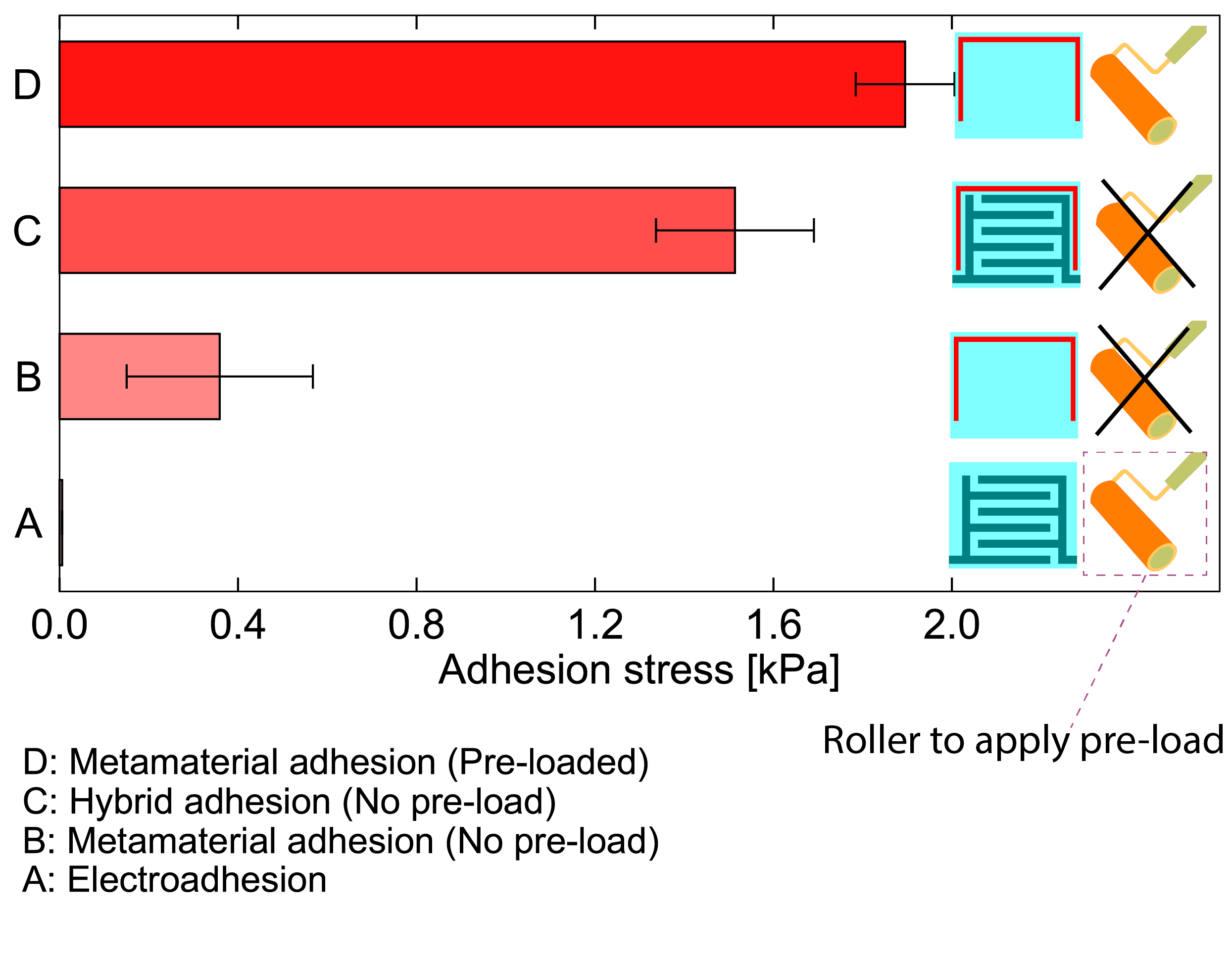}
\caption{Comparison with metamaterial adhesion, electroadhesion, and hybrid adhesion. A) Electroadhesion with preloading~\cite{caucciolo2022_peeling}. B) Metamaterial adhesion without preloading. C) Hybrid adhesion without preloading. D) Metamaterial adhesion with preloading.}
\label{fig:Literature_Comparison}
\end{figure}
\subsection{Metamaterial Simulation vs Experimental Results}
To validate the precision of the simulations, we conducted experiments on selected representative results from each set of internal points and subsequently compared these results with the simulations. The outcomes of this comparison are depicted in Fig.~\ref{fig:Sim_vs_Real}a. Our analysis reveals a good alignment between the experimental data and the simulation models, highlighting our capability to design and model metamaterial cuts for specific adhesion directionality requirements. The data also showcased maximum adhesive directionality with the 7-pt design, and good adhesion directionality with various 5-pt designs. However, for this work, we utilized the square-shaped design for the SEM adhesive. The first reason was to simplify the fitting of the electrode design within the shape geometry. The second reason was that we noticed that its forward adhesion stress was still the highest compared to the different designs, as seen in Fig.~\ref{fig:Sim_peeling}b. Thus, we decided to utilize this shape as a benchmark shape in this work and will pursue more intricate designs in the future, depending on the forward and reverse adhesion stress requirements. 
\subsection{Adhesion stress versus peeling distance}
Both types of samples were tested on a flat glass plate. For Fig.~\ref{fig:adhesion characterization}a, the metamaterial adhesive was preloaded with a roller, while for Fig.~\ref{fig:adhesion characterization}d, the SEM adhesive was not preloaded but activated with 4 kV$/$mm. From the results, we notice 4 peaks that resemble the peeling at each row of the adhesives when crack propagation occurs. The preloaded metamaterial adhesive had slightly higher adhesion stress in comparison to the SEM adhesive, 1.8 vs 1.6 kPa. For the metamaterial without roller preloading, the adhesion stress was accounted at 0.52 kPa, approximately 3.45x and 3.07x lower than either version, respectively. Thus, it highlights how well elecrostatic force attaches the adhesive to the object substrate.

\begin{figure}[b!]
\centering
\includegraphics[width=0.45\textwidth]{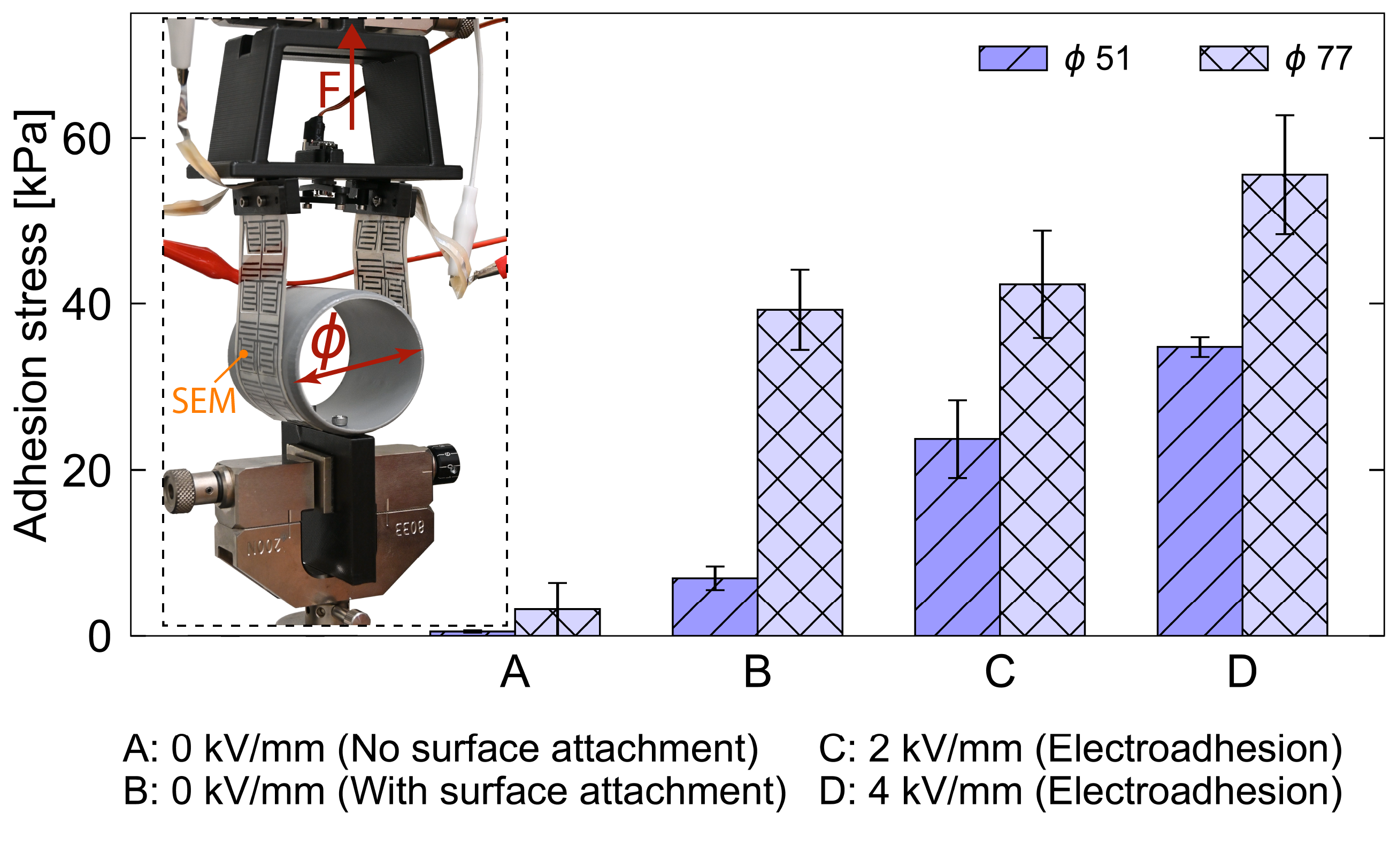}
\caption{Adhesion stress characterization of gripper on curved surfaces ($\phi$ 51 mm and 77 mm). Tips of adhesives are (A) not surface attached to the pipe, (B) surface attached. (C,D) Attached and with electric field of 2 kV$/$mm and 4 kV$/$mm is applied.}
\label{fig:Gripper_Characterize}
\end{figure}

\subsection{Adhesion directionality and enhancement}

The first directionality test was the comparison between the urethane versus the PDMS contact layer. As seen in  Fig.~\ref{fig:adhesion characterization}b and c, the enhancement ratio ($F_{\mathrm{max}/\mathrm{min}}$) on flat glass and wood surfaces are better for urethane, while the performance for curved surfaces is quite similar. For adhesion directionality, which means the ratio between forward and reverse peeling tests, are quite similar to each other, except for the flat glass and wood surfaces. Further, for manufacturing the adhesives, urethane layer had a faster curing time. Thus we proceeded to make the SEM adhesives with only Vytaflex 30, urethane contact layer. 

As we look at the different results for varying the voltage applied to the adhesive, in Fig.~\ref{fig:adhesion characterization}e and f, we notice the improvement in the forward peeling direction as the voltage increases. Even though 6 kV$/$mm was applicable for testing, the adhesives would break with prolonged and consistent testing. Thus, we focused on utilizing only 4 kV$/$mm. 

The enhancement ratio is also substantial as the voltage increases from 0-4 kV$/$mm but is more stable around 4-6 kV$/$mm, shown in Fig.~\ref{fig:adhesion characterization}f. The enhancement ratio was approximately 60x at 6 kV$/$mm for wood and PVC pipes, highlighting the multi-object adhesion capabilities of this system. The largest adhesion directionality value was $65.5x$ with the PVC pipe ($\phi$ 51 mm). In the case of enhancement ratio was $65.6x$. Overall, the adhesion directionality and enhancement ratio of the SEM adhesive have better properties than the metamaterial adhesive.

\begin{figure*}[t!]
\centering
\includegraphics[width=0.92\textwidth]{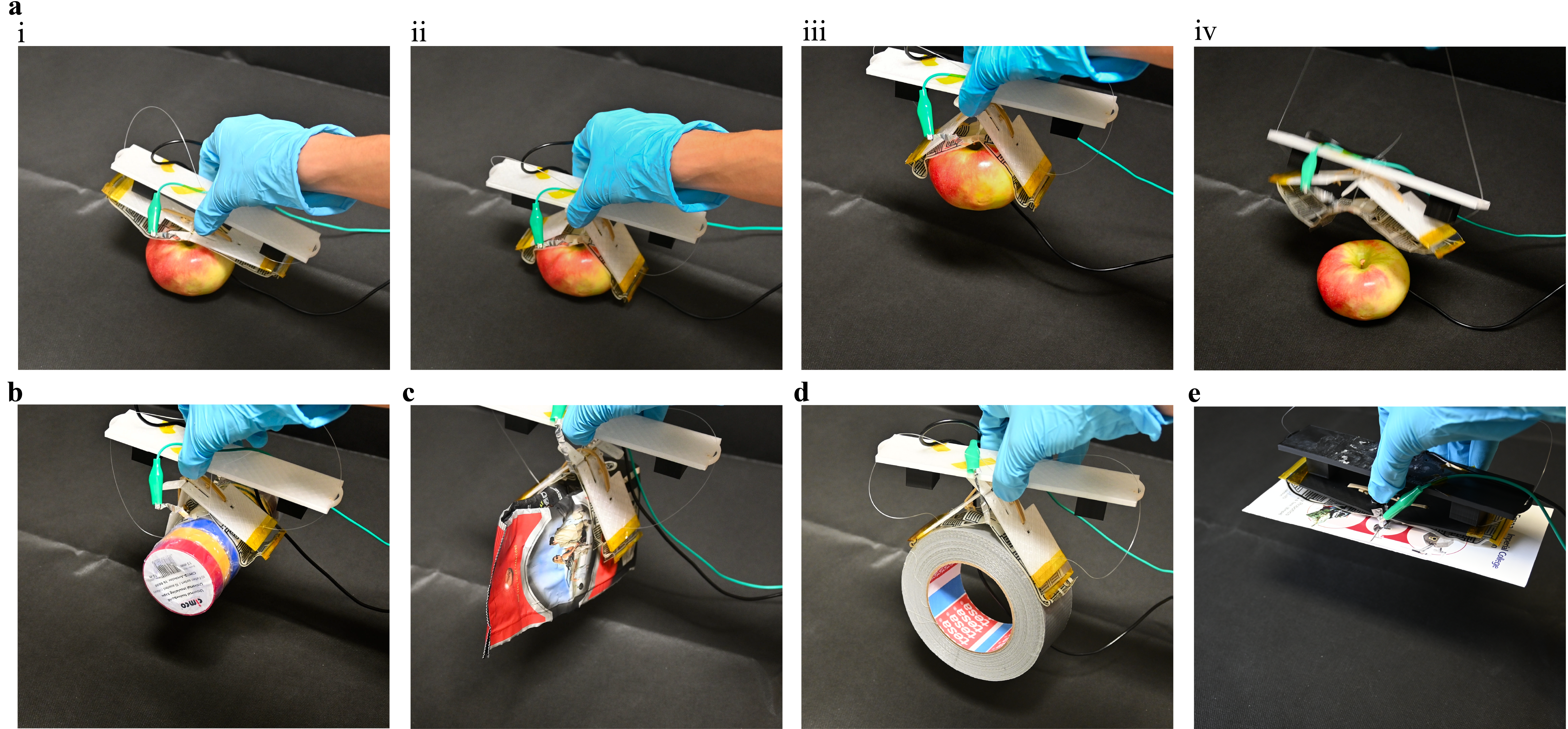}
\caption{SEM gripper experiment with a) an apple, sphere surface, that weighs 191 g (i: approaching the object, ii: bistable grasping with high voltage being applied, iii: lifting the object, iv: releasing the object). b) Tape (cylindrical shape, 284 g). c) Coffee bag (irregular shape, 350 g). d) Tape (circular surface, 499 g). e) Paper (flat surface, 10 g).}
\label{fig:demo}
\vspace{-1.5em}
\end{figure*}

\subsection{Adhesion comparison}

We compared the peeling force between the metamaterial~\cite{Hwang2023_metamaterial}, electrostatic adhesion~\cite{shintake2016versatile}, and SEM adhesives, as seen in Fig.~\ref{fig:Literature_Comparison}. For preloading, a standard rubber linoleum roller was utilized. We noticed that the electrostatic adhesive displayed quite low values even when activated at 6 kV$/$mm and that the preloaded metamaterial adhesion showed the highest adhesion stress. When not preloaded the metamaterial adhesion is weaker in comparison. To compare, the adhesion stress difference between the preloaded metamaterial and hybrid SEM adhesive was approximately 20\% less. The adhesive stress difference between SEM and metamaterial, without preloading, was approximately 76\% more.

\section{SEM Gripper Characterization}
\label{sec:gripper_charac}

In order to characterize the gripper adhesion shear force, a two-fingered grasping gripper (2.1g in total) was developed, as seen in Fig.~\ref{fig:Gripper_Characterize}. The two PVC pipes of $\phi$ 51 and $\phi$ 77 mm diameter were used to characterize this gripper, and the voltage was varied from 0-4 kV$/$mm. There were four testing conditions, A: Was without manually preloading and no voltage. B: Was preloading and no voltage. C and D were similar the difference was just the activation voltage, 2 and 4 kV$/$mm. Notice that the adhesion stress increased when applying higher voltages. The maximum stress was found at 55.6 kPa, which was approximately 33.3N pulling force on the $\phi$ 77 mm pipe. Thus, the gripper was able to lift up to $1617x$ of its own weight.

\section{Grasping Demonstration}
In order to demonstrate the developed SEM gripper for real-world applications, we performed pick-and-release demonstrations with a total of 5 different objects including deformable and delicate ones, with 4 kV$/$mm applied. The result can be seen in Fig.~\ref{fig:demo} and supplementary video. The objects ranged from an apple (191 g), a cylindrical tape roll (284 g), a coffee bag (350 g), a duct tape roll (500 g), and a flat surface pamphlet (10 g). From these tests, we highlighted both flat, curved, and spherical surface objects grasping and releasing at up to $237x$ the weight of the adhesives.

\section{Conclusion and Future Work}
\label{sec:conclude}
In this paper, we developed a Soft Electrostatic Metamaterial gripper. By combining the unique properties of metamaterials and electroadhesion, we successfully addressed the inherent limitations requiring the external preloading of metamaterial grippers. The SEM gripper demonstrated its ability for self-preloading and high adhesion force, allowing for pick-and-release actions of a variety of objects. Through our extensive characterization and testing, we unveiled the SEM gripper's proficiency in adhering to a multitude of surfaces, ranging from a flat glass pad to an irregularly shaped coffee bag. By modulating the applied voltage, we achieved a controlled adhesion force, enabling the gripper to lift objects up to $1617x$ of its weight. The pick-and-release demonstrations with objects of varied shapes and weights further exemplified the real-world applicability and versatility of the SEM gripper. Future research could delve into optimizing the metamaterial shape and electrode designs, further refining the gripper's sensitivity to handle even more delicate or intricately shaped objects. Further, developing automated grasping mechanisms would better classify the systems' capabilities for picking-and-placing. There also exists a future avenue to explore the integration of sensing capabilities for feedback control of the system, in order to verify safe attachment or not.

\section*{ACKNOWLEDGMENTS} 
We sincerely thank Prof. Dorina Opris and Ms. Jana Wolf for providing suggestions for dielectric fabrication methods. We also thank the members of the Laboratory of Sustainability Robotics and Aerial Robotics Lab for their support and stimulating discussions on this topic. 

This work was supported in part by EPSRC Awards (grant no, EP/R009953/1), in part by the Swiss National Science Foundation (SNSF) International Cooperation, EU ERANet grant (grant no, 194986), in part by the Royal Society Wolfson fellowship (grant no. RSWF/R1/18003), in part of the ERC Consolidator Grant as funded by (State Secretariat for Education, Research, and Innovation) SERI (grant no. MB22.00066), and in part by the Empa-EPFL-Imperial research partnership. 
\renewcommand*{\bibfont}{\footnotesize}
\printbibliography

\end{document}